\documentclass[notitlepage,12pt]{jedm}
\usepackage[sc,sf,small]{titlesec}
\usepackage[table]{xcolor}
\usepackage{url}
\usepackage{hyperref}
\usepackage{graphicx}
\hypersetup{
  colorlinks   = true, %Colours links instead of ugly boxes
  urlcolor     = blue, %Colour for external hyperlinks
  linkcolor    = blue, %Colour of internal links
  citecolor   = blue %Colour of citations
}

\begin{document}

\title{The Future of Learning in the Age of Generative AI: Automated Question Generation and Assessment with Large Language Models}
\date{} %do not delete this, it suppresses insertion of the date

%% Authors should be aligned.  If authors and their affiliation fit in a single horizontal block, you can follow this simple format:
\author{{\large Subhankar Maity}\\Department of Artificial Intelligence\\Indian Institute of Technology Kharagpur\\subhankar.ai@kgpian.iitkgp.ac.in \and {\large Aniket Deroy}\\Computer Science \& Engineering\\Indian Institute of Technology Kharagpur\\roydanik18@kgpian.iitkgp.ac.in}  

%% However, if authors do not fit in a single horizontal, they need to be put in parbox of fixed with with the \authorFixedWidth macro:
\newcommand{\authorFixedWidth}[1]{\parbox[t]{.25\textwidth}{\raggedright#1 \raisebox{0pt}[0pt][6pt]{}}}
%% The list of authors should look like:
%% \author{\authorFixedWidth{{\large Author One}\\Institution of Author one\\City, County\\email.of@author.dom} \and \authorFixedWidth{{\large Second Author}\\Affiliation\\City, County\\email@xx.xx}  \and \authorFixedWidth{{\large Third Author}\\Affiliation\\City, County\\email@xx.xx}   \and \authorFixedWidth{{\large Fourth Author}\\Affiliation\\City, County\\email@xx.xx}   \and \authorFixedWidth{{\large Fifth Author}\\A very long Affiliation for this author\\City, County\\email@xx.xx}   \and \authorFixedWidth{{\large Sixth Author}\\Affiliation\\City, County\\email@xx.xx}   \and \authorFixedWidth{{\large Seventh Author}\\Affiliation\\City, County\\email@xx.xx }}

\maketitle

\begin{abstract}
In recent years, large language models (LLMs) and generative AI have revolutionized natural language processing (NLP), offering unprecedented capabilities in education. This chapter explores the transformative potential of LLMs in automated question generation and answer assessment. It begins by examining the mechanisms behind LLMs, emphasizing their ability to comprehend and generate human-like text. The chapter then discusses methodologies for creating diverse, contextually relevant questions, enhancing learning through tailored, adaptive strategies. Key prompting techniques, such as zero-shot and chain-of-thought prompting, are evaluated for their effectiveness in generating high-quality questions, including open-ended and multiple-choice formats in various languages. Advanced NLP methods like fine-tuning and prompt-tuning are explored for their role in generating task-specific questions, despite associated costs. The chapter also covers the human evaluation of generated questions, highlighting quality variations across different methods and areas for improvement. Furthermore, it delves into automated answer assessment, demonstrating how LLMs can accurately evaluate responses, provide constructive feedback, and identify nuanced understanding or misconceptions. Examples illustrate both successful assessments and areas needing improvement. The discussion underscores the potential of LLMs to replace costly, time-consuming human assessments when appropriately guided, showcasing their advanced understanding and reasoning capabilities in streamlining educational processes. \\ %Keep \\ for spacing to keywords

{\parindent0pt
\textbf{Keywords:} Natural Language Processing (NLP), Large Language Models (LLMs), Education, Automated Question Generation (AQG), Answer Assessment, Prompt Engineering
}
\end{abstract}

\section{Introduction}

The educational landscape is evolving rapidly, driven by the integration of advanced technologies that challenge traditional teaching methods. Among these technologies, Large Language Models (LLMs) have emerged as powerful tools, capable of revolutionizing the way we approach learning and assessment. These models, epitomized by systems such as GPT-4 \cite{a1} and beyond, have demonstrated an extraordinary ability to understand and generate human-like text, enabling them to perform tasks that were once the exclusive domain of human educators \cite{a2,a3}. In the realm of education, question generation and assessment are critical components that shape the learning experience. Traditionally, these tasks require significant human effort, involving educators in the meticulous design of questions that not only test knowledge but also promote deeper understanding \cite{a4}. Assessing student responses, particularly in open-ended formats, is another labor-intensive task that demands careful consideration of context, nuance, and individual student needs \cite{a5}. However, as the demand for personalized and adaptive learning grows, the limitations of human-driven approaches have become more apparent.

This chapter delves into the transformative potential of LLMs in automating these crucial educational tasks. We explore how LLMs can be leveraged to generate a wide variety of questions—ranging from simple factual queries to complex, open-ended questions—that are contextually relevant and aligned with educational goals \cite{a6,a7,a8}. We also examine the capabilities of LLMs in automated answer assessment, where these models can evaluate student responses, offer feedback, and even identify subtle misconceptions, all at a scale and efficiency that human educators cannot match \cite{a11}. The introduction of LLMs into the educational process is not without challenges. Issues such as the quality and relevance of generated questions, the accuracy of automated assessments, and the ethical implications of relying on AI for education require careful consideration \cite{a12}. 

This chapter addresses these concerns, offering insights into how LLMs can be guided and refined to ensure they complement and enhance human-led education rather than replacing it. In the sections that follow, we will first provide a detailed overview of LLMs, focusing on their architecture and underlying mechanisms. This will set the stage for a discussion on various methodologies and prompting techniques used to generate educational questions. We will then explore the role of advanced NLP methods such as fine-tuning and prompt-tuning in enhancing the quality and specificity of generated questions. The chapter will also cover human evaluation metrics for assessing the quality of these questions and the performance of LLMs in automated answer assessment. Finally, we will discuss the broader implications of integrating LLMs into education, highlighting both their potential benefits and the challenges that must be addressed to fully realize their capabilities.

%%%%%%%%%%%%%%%%%%%%%%%%%%%%%%%%%%%%%%%%%%%%%%%%%%%%%%%%%%%%%%%%%%%%%%%%%%%%%
\section{Understanding Large Language Models in Education}

\subsection{The Architecture and Mechanisms of LLMs}

Large Language Models (LLMs), built on the foundations of deep learning and transformer architectures \cite{a13}, have brought about a paradigm shift in natural language processing (NLP). These models, trained on vast corpora of text data, are designed to predict and generate text based on a given input \cite{a14}. Their ability to understand context, recognize patterns, and generate coherent, contextually appropriate text makes them particularly well-suited for educational applications.

At the core of LLMs is the transformer architecture, which uses self-attention mechanisms to weigh the importance of different words in a sentence relative to each other \cite{a13}. This allows the model to capture long-range dependencies in text, making it capable of understanding complex sentences and generating nuanced responses. For educational purposes, this means LLMs can generate questions that are not only grammatically correct but also contextually relevant and pedagogically sound. The training process of LLMs involves exposure to diverse datasets that cover a wide range of topics and writing styles \cite{a15}. This extensive training enables the models to develop a broad understanding of language, which they can then apply to specific tasks such as question generation and assessment. However, while LLMs excel in generating human-like text, their effectiveness in educational contexts depends on how well they are guided and fine-tuned for specific tasks.

\subsection{The Role of Fine-Tuning and Prompt-Tuning}

To adapt LLMs for educational question generation and assessment, techniques such as fine-tuning and prompt-tuning are employed. Fine-tuning involves training the LLM on a specialized dataset that is closely aligned with the target task. This allows the model to learn the nuances of educational content and generate questions that are more closely aligned with the curriculum and learning objectives \cite{a16}.

Prompt-tuning, on the other hand, involves designing specific prompts that guide the LLM in generating the desired output \cite{a17}. This technique leverages the model's existing knowledge and directs it towards generating contextually relevant and pedagogically valuable questions. For instance, a prompt might instruct the LLM to generate a question based on a specific passage of text, encouraging the model to focus on key concepts and ideas that are essential for learning.

Both fine-tuning and prompt-tuning have their advantages and challenges. Fine-tuning can produce highly specialized models that excel in specific tasks, but it is resource-intensive and requires access to large, high-quality datasets \cite{a18}. Prompt-tuning, while more flexible and less resource-demanding, relies heavily on the design of effective prompts and may not always achieve the same level of specificity as fine-tuned models \cite{a17}. Despite these challenges, both techniques have shown significant promise in enhancing the performance of LLMs in educational settings.

\section{Automated Question Generation: Methodologies and Techniques}

\subsection{Generating Diverse and Contextually Relevant Questions}

The automated generation of questions using large language models (LLMs) represents a powerful tool in education, enabling the creation of diverse and contextually relevant questions tailored to various learning objectives \cite{a7}. The methodologies employed in question generation are varied, each contributing to the quality and applicability of the generated content. Below are the key methods utilized in this domain:

\begin{itemize}
    
\item \textbf{Zero-Shot Prompting}: Zero-shot learning allows models like GPT-3 \cite{a2} to generate questions based on minimal instructions. The model leverages its pre-trained knowledge to generate relevant questions directly from the provided text, without the need for additional examples or fine-tuning \cite{a2}. This approach is particularly useful for generating questions across a wide range of topics, but the quality may vary depending on the complexity of the input text \cite{a6,a10}.

\item \textbf{Few-Shot Prompting}: Few-shot prompting provides the model with a few examples of the task to guide its question generation. By including a few question-answer pairs as part of the prompt, this method enhances the model's understanding of the task, leading to improved relevance and quality of the generated questions \cite{a2}. This technique is effective in scenarios where the desired question format or content is more complex and needs to be clearly defined for the model.

\item \textbf{Chain-of-Thought Prompting}: A structured technique that involves guiding the LLM through a step-by-step reasoning process before it generates the final question. For example, the model may first be asked to summarize a passage, identify key concepts, and then generate a question that tests understanding of these concepts \cite{a19,a9}. This approach is particularly effective for generating higher-order questions that require critical thinking and analysis, ensuring that the questions align with specific educational goals.

\item \textbf{Fine-Tuning}: Fine-tuning involves further training the LLM on a specific dataset of questions and answers relevant to the target domain. By learning the patterns and structures of effective questions from the training data, fine-tuning allows the model to generate more accurate and context-specific questions \cite{a18}. This method is resource-intensive but results in highly specialized models that can produce high-quality questions tailored to specific subjects or curricula \cite{a6}.

\item \textbf{Prompt-Tuning}: A recent and computationally efficient technique, prompt-tuning involves adjusting a small set of parameters (the prompt) while leaving the rest of the model unchanged. This method has proven effective in generating high-quality questions across various educational contexts, especially when the goal is to adapt a general-purpose LLM to a specific task without extensive retraining \cite{a17}. Prompt-tuning allows for quick adaptation and customization of LLMs to generate questions that are both relevant and aligned with specific educational objectives.

\item \textbf{Multiformat and Multilingual Question Generation}: LLMs are capable of generating both open-ended \cite{a6} and multiple-choice questions \cite{a9}, catering to different assessment needs. Open-ended questions encourage critical thinking and exploration, while multiple-choice questions are useful for evaluating specific knowledge or skills \cite{a9}. Additionally, the multilingual capabilities of LLMs enable the generation of questions in various languages, making them valuable tools for language learning and cross-cultural education \cite{a14,a9}.

\end{itemize}

These methodologies, when applied effectively, enhance the educational process by generating diverse, high-quality questions that cater to different learning contexts and objectives. As LLMs continue to evolve, the integration of these techniques will further improve the relevance, accuracy, and utility of automated question generation in education.

\subsection{Types of Questions Generated by LLMs}

In the context of education, different types of questions serve varied pedagogical functions, and LLMs are capable of generating a broad spectrum of question types. Below are the primary categories:

\begin{itemize}
   
\item \textbf{Factual Questions}: These questions focus on the recall of specific information, such as dates, definitions, or events. They are typically straightforward and aim to assess the student’s memory and basic understanding of the subject matter \cite{a20}. 

Example: "\textit{What is the capital of France?}"

\item \textbf{Open-Ended Questions}: Open-ended questions are designed to encourage deep thinking and exploration, allowing students to express their thoughts freely and creatively. These questions do not have a single correct answer, promoting critical thinking and discussion \cite{a20,a6}. 

Example: "\textit{What does purchasing power parity do?}"

\item \textbf{Multiple-Choice Questions (MCQs)}: MCQs assess specific knowledge or skills by providing a set of possible answers from which the student must choose the correct one. They are widely used for their efficiency in testing and grading \cite{a9}.

Example: "\textit{Which of the following is the largest planet in our solar system?}

(a) \textit{Earth} (b) \textit{Jupiter} (c) \textit{Mars} (d) \textit{Venus}"
 
\end{itemize}

LLMs, through their sophisticated language processing capabilities, can generate these varied question types effectively, adapting them to different educational contexts and learning objectives.

\section{Automated Answer Assessment: Evaluating Student Responses}

\subsection{The Capabilities of LLMs in Automated Answer Assessment}

In addition to generating questions, LLMs have demonstrated significant potential in automated answer assessment \cite{a11}. The ability to accurately evaluate student responses and provide feedback is a critical component of the educational process \cite{a11}. Traditionally, this task has been performed by human educators, who must carefully consider the content, context, and nuance of each response \cite{a21}. However, as the demand for personalized and scalable education grows, the limitations of human-driven assessment become more apparent \cite{a22}.

LLMs offer a scalable solution to automated answer assessment, with the ability to evaluate a wide range of responses, from simple factual answers to complex, open-ended essays  \cite{a11}. By leveraging their deep understanding of language and context, LLMs can identify key concepts, assess the accuracy of the response, and provide constructive feedback \cite{a23}. This capability is particularly valuable in large-scale educational settings, where the volume of student responses can be overwhelming for human assessors \cite{a24}.

One of the key strengths of LLMs in automated assessment is their ability to identify nuanced understanding or misconceptions in student responses \cite{a25}. For example, an LLM can evaluate an essay on a historical event, recognizing whether the student has grasped the underlying causes and implications of the event, rather than simply recounting facts \cite{a26}. 

However, while LLMs have shown great promise in automated assessment, there are challenges to be addressed \cite{a11}. One of the primary concerns is the accuracy and consistency of the assessments. LLMs, like all AI systems, are not infallible and can sometimes produce incorrect or biased evaluations \cite{a27}. Ensuring that the assessments are fair, accurate, and aligned with the learning objectives is crucial for the successful integration of LLMs into the educational process \cite{a11}.

\subsection{Examples of Successful Assessments and Areas for Improvement}

To illustrate the capabilities of LLMs in automated answer assessment, consider the following examples:

\begin{itemize}
    
\item \textbf{Short-Answer Evaluation}: An LLM is tasked with evaluating short-answer responses in a biology exam \cite{a28}. The model is able to accurately assess whether the student has correctly identified the function of a specific organelle within a cell, providing feedback on both correct and incorrect answers. The LLM also identifies common misconceptions, such as confusing the roles of the mitochondria and the nucleus, and provides corrective feedback to guide the student's learning.

\item \textbf{Essay Grading}: In a history class, students are asked to write essays on the causes and effects of World War II. The LLM evaluates the essays based on criteria such as understanding of key events, analysis of historical factors, and coherence of argument. The model is able to identify well-reasoned arguments and provide feedback on areas where the student could improve, such as providing more evidence or considering alternative perspectives \cite{a29,a30}.

\item \textbf{Multiple-Choice Question Analysis}: An LLM is used to analyze student responses to multiple-choice questions in a mathematics exam \cite{a30}. In addition to identifying the correct answers, the model also analyzes the patterns of incorrect responses, identifying common errors and misconceptions. This information is used to provide targeted feedback and suggest areas for further study.

\end{itemize}
While these examples demonstrate the potential of LLMs in automated assessment, there are also areas for improvement. One challenge is ensuring that the feedback provided by the LLM is constructive and actionable \cite{a31}. For instance, while the model may correctly identify an error in a student's response, it must also provide clear guidance on how to address the mistake. Additionally, the LLM must be able to adapt its feedback to the individual needs of each student, taking into account their prior knowledge and learning style.

Another area for improvement is the ability of LLMs to assess more complex and creative responses, such as those involving critical thinking, problem-solving, or artistic expression. While LLMs have made significant strides in understanding and generating text, evaluating these higher-order skills remains a challenge \cite{a32}. Future research and development will be needed to enhance the capabilities of LLMs in these areas, ensuring that they can fully support the diverse needs of learners.

\section{Human Evaluation and Quality Metrics for Generated Questions}

\subsection{Assessing the Quality of Generated Questions}

The quality of questions generated by LLMs is a critical factor in their effectiveness as educational tools. High-quality questions should be clear, relevant, and aligned with the learning objectives, challenging students to think critically and apply their knowledge. To ensure that the questions generated by LLMs meet these standards, human evaluation and quality metrics play a crucial role \cite{a33}.

Human evaluation involves assessing the generated questions based on a set of predefined criteria, such as grammaticality, relevance, clarity, complexity, and alignment with the curriculum \cite{a33,a6}. Expert educators or subject matter experts typically conduct this evaluation, providing feedback on the strengths and weaknesses of the questions. This feedback is invaluable for refining the prompts and improving the quality of the generated questions.

In addition to human evaluation, automated quality metrics can be used to assess the generated questions. These metrics may include measures such as unigram-, bigram-, and n-gram-based evaluations, which provide quantitative insights into the quality of the questions \cite{a33}. However, these automated evaluation metrics used for assessing LLM-generated questions have limitations. This limitation arises because these metrics often prioritize linguistic similarity (e.g., character, unigrams, bigrams, or longest common subsequence-based overlap) rather than deeper contextual understanding \cite{a34}.

One of the challenges in evaluating the quality of generated questions is the subjective nature of some of the criteria. For instance, what one educator considers a challenging and thought-provoking question, another might view as overly complex or unclear \cite{a35}. To address this, it is important to establish clear guidelines and criteria for evaluation, ensuring consistency and objectivity in the assessment process.

\subsection{Variations in Quality Across Different Methods}

The quality of questions generated by LLMs can vary significantly depending on the methods and techniques used. For example, questions generated using zero-shot prompting may be more general and less tailored to the specific content, while those generated using fine-tuning or prompt-tuning may be more precise and relevant \cite{a6}. Understanding these variations is essential for selecting the appropriate method for a given educational context.

One common variation in quality is related to the complexity of the generated questions. LLMs are capable of generating both simple, factual questions and more complex, analytical questions \cite{a10}. However, the latter requires a deeper understanding of the content and context, which may not always be achievable through basic prompting techniques. To generate higher-order questions, more advanced techniques, such as chain-of-thought prompting \cite{a19} or fine-tuning \cite{a18}, may be necessary.

Another variation in quality is related to the cultural and linguistic diversity of the generated questions. LLMs trained on diverse datasets are better equipped to generate questions that are culturally relevant and appropriate for different student populations. However, this diversity can also introduce challenges, as the model may generate questions that are less familiar or relevant to certain groups of students. Ensuring that the generated questions are inclusive and accessible to all learners is an important consideration in the evaluation process \cite{a7,a10}.

\section{Broader Implications and Future Directions}

\subsection{The Role of LLMs in Personalized and Adaptive Learning}

As LLMs continue to evolve, their role in personalized and adaptive learning is becoming increasingly significant. The ability of LLMs to generate contextually relevant questions and assess student responses on a large scale opens up new possibilities for personalized education \cite{a36}. By leveraging LLMs, educators can create tailored learning experiences that adapt to the individual needs and progress of each student \cite{a37}.

One of the key benefits of using LLMs in personalized learning is the ability to provide immediate feedback and guidance \cite{a38}. As students interact with the system, LLMs can generate questions that challenge their understanding, identify areas of difficulty, and offer targeted feedback to support their learning. This real-time interaction can help students stay engaged and motivated, while also providing educators with valuable insights into their progress.

However, the integration of LLMs into personalized learning also raises important questions about the balance between human and AI-driven education \cite{a39}. While LLMs can offer scalable and efficient solutions, they cannot replace the nuanced understanding and empathy that human educators bring to the classroom. The challenge lies in finding the right balance, where LLMs complement and enhance human-led education, rather than supplanting it.

\subsection{Ethical Considerations and Challenges}

The use of LLMs in education also raises important ethical considerations \cite{a38}. Issues such as bias, fairness, and transparency are central to the responsible use of AI in education \cite{a40}. LLMs, like all AI systems, are trained on data that may contain biases, and these biases can be reflected in the questions they generate or the assessments they perform \cite{a40}. Ensuring that LLMs are fair and unbiased requires careful attention to the training data, as well as ongoing monitoring and evaluation of the system's outputs.

Another ethical consideration is the transparency of the AI-driven educational process \cite{a41}. Students and educators need to understand how LLMs generate questions and assess responses, and they should be informed about the potential limitations and biases of the system \cite{a40}. Transparency is key to building trust in AI-driven education and ensuring that students and educators feel confident in the use of these technologies \cite{a42}.

Finally, the use of LLMs in education raises questions about data privacy and security \cite{a43}. As LLMs interact with students and assess their responses, they may collect and store sensitive information about the student's performance and learning history. Protecting this data and ensuring that it is used responsibly is essential for maintaining the integrity and security of the educational process.

\subsection{Future Directions in Automated Question Generation and Assessment}

Looking to the future, the role of LLMs in automated question generation and assessment is likely to expand and evolve \cite{a11}. Advances in AI and NLP technologies will enable the development of more sophisticated models that are better equipped to handle complex and creative educational tasks \cite{a44}. As these models become more integrated into the educational process, they will play a key role in supporting personalized and adaptive learning, providing scalable solutions that enhance the quality and accessibility of education.

One promising direction for future research is the development of models that can assess higher-order thinking skills, such as critical thinking, problem-solving, and creativity. These skills are essential for success in the 21st century, and the ability to assess them accurately and efficiently is a major challenge for educators. LLMs, with their advanced language understanding and generation capabilities, have the potential to address this challenge, providing new tools for assessing and supporting the development of these critical skills \cite{a45}.

Another important direction for future research is the exploration of new methodologies for fine-tuning and prompt-tuning LLMs for specific educational tasks. As LLMs continue to be used in a wider range of educational contexts, it will be important to develop techniques that allow for the efficient and effective adaptation of these models to different subject areas, student populations, and learning objectives.

\section{Conclusion}

In conclusion, large language models have the potential to revolutionize education through automated question generation and answer assessment. These models, with their ability to understand and generate human-like text, offer scalable solutions that can enhance personalized and adaptive learning. By leveraging advanced prompting techniques and fine-tuning methodologies, educators can create high-quality, contextually relevant questions that challenge students and support their learning. Furthermore, LLMs' capabilities in automated assessment can provide timely and constructive feedback, helping students identify areas for improvement and guiding their educational journey.

However, the integration of LLMs into education also presents challenges and ethical considerations that must be carefully addressed. Ensuring the fairness, accuracy, and transparency of AI-driven educational processes is essential for building trust and confidence in these technologies. As we look to the future, ongoing research and development will be key to realizing the full potential of LLMs in education, creating a more personalized, adaptive, and accessible learning experience for all students.

% REMOVE NOCITE OR IT WILL LIST EVERYTHING IN YOUR DATABASE AS A REFERENCE
%\nocite{*}

\bibliographystyle{acmtrans}
\bibliography{ref}

\end{document}